\documentclass[11pt]{article}

\usepackage{acl}

\usepackage{times}
\usepackage{latexsym}
\usepackage{booktabs}
\usepackage{multirow}
\usepackage{adjustbox}
\usepackage{placeins}
\usepackage{xurl}
\usepackage{hyperref}
\usepackage[table]{xcolor}
\usepackage{todonotes} 
\usepackage{float}

\definecolor{TAPOBlue}{RGB}{238,246,252}
\definecolor{blue1}{RGB}{228,242,252}
\definecolor{DropRed}{RGB}{213,32,52}
\definecolor{green1}{RGB}{0,137,72}

\usepackage[most]{tcolorbox}
\usepackage{xcolor}
\usepackage{enumitem}

\definecolor{boxgray}{RGB}{95,95,95}
\definecolor{bordergray}{RGB}{90,90,90}

\newtcolorbox{appendixpromptbox}[1]{
  enhanced,
  breakable,
  colback=white,
  colframe=bordergray,
  boxrule=0.8pt,
  arc=2pt,
  left=7pt,
  right=7pt,
  top=8pt,
  bottom=7pt,
  fontupper=\small,
  attach boxed title to top left={
    xshift=6pt,
    yshift=-2pt
  },
  boxed title style={
    colback=boxgray,
    colframe=boxgray,
    arc=2pt,
    boxrule=0pt,
    left=7pt,
    right=7pt,
    top=2pt,
    bottom=2pt
  },
  title={#1},
  fonttitle=\bfseries\small\color{white}
}

\usepackage[T1]{fontenc}
\usepackage[utf8]{inputenc}

\usepackage{microtype}

\usepackage{inconsolata}

\usepackage{graphicx}
\usepackage{amsmath}
\title{Trait-Aware Policy Optimization \\ for Autoregressive Multi-Trait Essay Scoring}

\author{
\textbf{Zhengyang Wang\textsuperscript{1,2}} \quad
\textbf{Sanwoo Lee\textsuperscript{1,3}} \quad
\textbf{Jiaxin Wang\textsuperscript{1,3}} \\
\textbf{Chenxi Miao\textsuperscript{4}} \quad
\textbf{Weikang Li\textsuperscript{4}} \quad
\textbf{Yunfang Wu\textsuperscript{1,3}\thanks{Corresponding author.}} \\
\textsuperscript{1}National Key Laboratory for Multimedia Information Processing, Peking University \\
\textsuperscript{2}School of Software and Microelectronics, Peking University \\
\textsuperscript{3}School of Computer Science, Peking University \\
\textsuperscript{4}Baidu Inc. \\
\texttt{zhengyangwang25@stu.pku.edu.cn},
\texttt{wuyf@pku.edu.cn}
}

\begin{document}
\maketitle

\begin{abstract}
Multi-trait essay scoring aims to provide fine-grained evaluation of writing quality across multiple dimensions. 
However, how to effectively post-train autoregressive scoring models remains underexplored. In this paper, we propose Trait-Aware Policy Optimization (TAPO), a post-training framework tailored to autoregressive multi-trait scoring. Our method decomposes rewards along both the sample and trait dimensions, combining global scoring consistency, trait-level accuracy, format validity, and inter-trait dependency preservation. In addition, we use enhanced prompts throughout training by incorporating original prompt texts and trait descriptions, providing richer semantic information for trait-specific score generation. Experiments across multiple backbone models show that our method consistently improves multi-trait scoring performance over supervised fine-tuning and scalar-reward optimization baselines, demonstrating the effectiveness and transferability of trait-aware post-training for essay scoring.
\end{abstract}

\section{Introduction}

\begin{figure}[t] \includegraphics[width=\columnwidth]{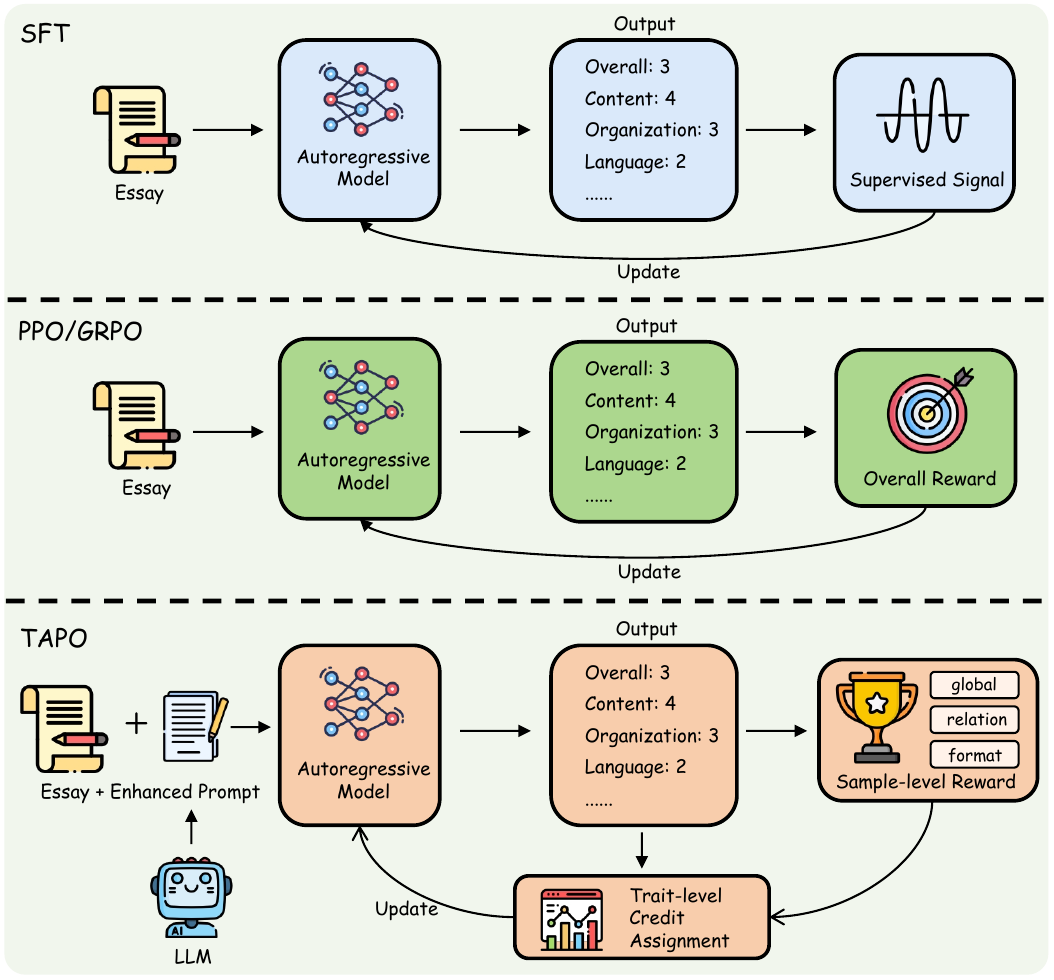} \caption{Overview of SFT, PPO, GRPO, and TAPO for autoregressive multi-trait essay scoring.} 
\label{fig:figure1} \end{figure}
%\todo[linecolor=red]{这里的图要改吗}没想好怎么改，暂时保持原状

Automated essay scoring (AES) has long been studied as a way to reduce the cost of large-scale writing assessment. Most neural AES models formulate scoring as a classification or regression problem, predicting a single holistic score for each essay. While holistic scoring provides a compact assessment of overall writing quality, it offers limited diagnostic feedback and does not reveal which aspects of writing contribute to the final score.

Multi-trait essay scoring addresses this limitation by predicting scores for multiple writing dimensions, such as content, organization, word choice, sentence fluency, and conventions. This setting is more informative, but also more challenging: a model must identify evidence relevant to each trait while preserving dependencies among correlated writing dimensions. Existing methods have improved multi-trait AES by modeling trait dependencies through shared representations and structured interactions \citep{kumarManyHandsMake2022}. Recent studies have reformulated this task as autoregressive score generation, enabling language models to produce structured trait scores instead of relying on conventional classification or regression heads. 
%or autoregressive score generation. 
In particular, ArTS reformulates multi-trait AES as structured score generation \citep{doAutoregressiveScoreGeneration2024}, and SaMRL further introduces PPO-style reinforcement learning with scoring-aware rewards \citep{doAutoregressiveMultitraitEssay2024}.

%advances in LLM post-training have introduced 
Group Relative Policy Optimization (GRPO)
%, which 
has shown strong effectiveness in reinforcement learning with verifiable rewards (RLVR) \citep{shaoDeepSeekMathPushingLimits2024}. Compared with PPO-style optimization, GRPO does not require an explicit value model and estimates advantages by comparing multiple sampled responses for the same input \citep{schulmanProximalPolicyOptimization2017,shaoDeepSeekMathPushingLimits2024}. %These properties make GRPO appealing for autoregressive multi-trait 
In the AES task, once a structured output is parsed, its quality can be automatically evaluated. Therefore, multi-trait AES naturally provides verifiable reward signals, making GRPO an efficient and metric-aware post-training framework for this task.

However, the standard GRPO provides only a sequence-level scalar reward. In multi-trait AES, a generated response contains multiple trait-specific scoring decisions, where some traits may be correct while others are incorrect. %Such 
The conventional coarse feedback cannot fully exploit trait-level information during post-training, motivating our trait-aware adaptation of GRPO. 
%This mismatch motivates a trait-aware adaptation of GRPO.

Another challenge lies in prompt and trait understanding. %Autoregressive language models are sensitive to semantic cues due to large-scale pretraining, but
Previous autoregressive language-model work uses only a prompt ID and
trait names \citep{doAutoregressiveScoreGeneration2024}, which provides limited semantic information about the writing task and scoring dimensions. Recent LLM-based approaches, such as RMTS \citep{chuRationaleEssayScores2025} and RaDME~\citep{doTeachtoreasonScoringSelfexplainable2026}, improve multi-trait scoring by using LLM-generated %evaluations or 
rationales to transfer rubric-level knowledge to smaller 
%scoring 
models. In contrast, we use an input-side enhancement that does not require generating or supervising additional rationales: we augment the prompt ID with the original prompt text and append
natural-language descriptions of the scoring traits throughout training, providing explicit semantic guidance for score generation. %without relying on full rationale distillation.

To this end, we propose \textbf{Trait-Aware Policy Optimization} (TAPO), a trait-aware post-training framework for autoregressive multi-trait essay scoring. Figure~\ref{fig:figure1} compares TAPO with existing autoregressive training paradigms. TAPO adapts reward design and advantage computation to structured multi-trait outputs by combining sample-level scoring signals with local trait-level feedback, enabling overall score optimization while assigning localized credit to individual trait predictions.

Experiments using T5-large and Qwen3-1.7B on ASAP and ASAP++, as well as
Feedback Prize, show that our approach achieves the strongest average
performance among the compared methods under trait-wise and prompt-wise
evaluations.
%Further cross-backbone, cross-dataset, and ablation studies demonstrate its generalization ability and validate the effectiveness of its core components.

Our contributions are summarized as follows:
\begin{itemize}
    \item We propose TAPO, a post-training framework that adapts policy optimization to structured multi-trait scoring through multi-dimensional rewards and trait-level credit assignment.
    \item We introduce enhanced prompts with original prompt texts and LLM-generated trait descriptions to provide explicit semantic guidance for trait-specific score generation.
    \item Experiments show that our approach achieves strong performance on
    multi-trait essay scoring benchmarks, with consistent improvements across
    T5-large and Qwen3-1.7B and across ASAP/ASAP++ and Feedback Prize.
\end{itemize}

\section{Related Work}
\label{sec:related_work}

\paragraph{Automated Essay Scoring.}
Early AES research mainly relied on handcrafted linguistic, lexical, semantic, and discourse features, as seen in LSA-based assessment and e-rater-style scoring systems \citep{foltzIntelligentEssayAssessor1999,landauerAutomaticEssayAssessment2003,attaliAutomatedEssayScoring2006}. Later neural approaches learned essay representations with recurrent networks, attention-based models, string kernels with embeddings, and pretrained language models \citep{cozmaAutomatedEssayScoring2018,alikaniotisAutomaticTextScoring2016,taghipourNeuralApproachAutomated2016,dongAttentionbasedRecurrentConvolutional2017,utoNeuralAutomatedEssay2020,wangUseBertAutomated2022,boquioCanonicalFinetuningLeveraging2024,dasTransformerbasedJointModelling2024,yangEnhancingAutomatedEssay2020}. However, most of these methods focus on holistic scoring and provide limited diagnostic feedback for specific writing traits.

\paragraph{Multi-Trait Essay Scoring.}
Multi-trait essay scoring extends AES by predicting scores for multiple writing dimensions. The ASAP++ dataset enriches the original ASAP benchmark with trait-level annotations, enabling systematic evaluation of analytic essay scoring models \citep{mathiasASAPEnrichingASAP2018}. Early studies adopt multi-task learning to share essay representations across holistic and trait-level scoring tasks while using trait-specific prediction components \citep{mathiasCanNeuralNetworks2020,kumarManyHandsMake2022,DualscaleBERTUsing2024,doPromptTraitRelationaware2023}. Later methods move beyond shared representations by explicitly modeling trait dependencies and prompt-specific criteria, using cross-prompt modeling, mixture-of-experts architectures, graph-based interactions, grammar-aware representations, or multi-knowledge enhancement \citep{ridleyAutomatedCrosspromptScoring2021,wangTMESTraitAwareMixofExperts2025,liGraphBasedMultiTraitEssay2025,doPromptGeneralizationGrammaraware2025,zhaoMultiknowledgeEnhancedGraph2026}. Another line of work shifts from regression or classification to autoregressive score generation. ArTS redefines multi-trait AES as a score-generation task with T5, while SaMRL further introduces reinforcement learning with QWK-based rewards and MSE penalties \citep{doAutoregressiveScoreGeneration2024,doAutoregressiveMultitraitEssay2024}. Recent LLM-based methods further improve trait understanding by generating scoring criteria, rubric-assisted features, rationales, self-explanations, or trait-description-aware representations \citep{leeUnleashingLargeLanguage2024,eltanboulyTRATESTraitSpecificRubricAssisted2025,chuRationaleEssayScores2025,doTeachtoreasonScoringSelfexplainable2026,faveroHolisticScoresAutomatic2026,choiEnhancingAutomatedEssay2026,sunUniMTESUnifiedFramework2026}. However, existing autoregressive approaches still mainly rely on sample-level or sequence-level feedback, leaving trait-level information insufficiently exploited during post-training.

%这里补充了对世航文章的引用和对比
Concurrent work explores joint feedback and score generation through teacher-distilled rubric-grounded explanations \citep{yang-etal-2026-hifts}. 
In contrast, our work focuses on trait-localized credit assignment for structured score generation, using trait-level prediction errors to provide localized reinforcement signals for each generated score.

\paragraph{Reinforcement Learning for Generative Models.}
Reinforcement learning has become an important tool for post-training generative language models. PPO improves training stability with a clipped policy optimization objective and has been widely used in RLHF \citep{schulmanProximalPolicyOptimization2017,zieglerFineTuningLanguageModels2019,ouyangTrainingLanguageModels2022,tuanProximalPolicyOptimization2018}. GRPO further reduces the cost of PPO-style optimization by estimating advantages from a group of sampled responses, avoiding the need for an explicit value model \citep{shaoDeepSeekMathPushingLimits2024}. Recent work improves GRPO-style optimization through dynamic sampling, completion pruning, sequence-level policy optimization, optimization-bias correction, and value-augmented optimization \citep{yuDAPOOpenSourceLLM2025,linCPPOAcceleratingTraining2025,zhengGroupSequencePolicy2025,liuUnderstandingR1ZeroLikeTraining2025,yueVAPOEfficientReliable2025}. Fine-grained credit assignment has also attracted increasing attention, with methods deriving token- or step-level signals from process rewards, distribution-guided reallocation, eligibility traces, generated critiques, or outcome-grounded advantage reshaping \citep{uesatoSolvingMathWord2022,LetsVerifyStep,jinDGPODistributionGuided2026,parthasarathiGRPO$lambda$CreditAssignment2025,xieCAPOEnhancingLLM2025,liOutcomeGroundedAdvantageReshaping2026}. Unlike these methods, our work focuses on structured multi-trait essay scoring, where each generated score corresponds to a known writing trait, enabling localized feedback from trait-level prediction errors.

\section{Preliminaries}
\subsection{Problem Formulation} 
\label{sec:problem_formulation}

% In this paper, we distinguish a \textit{prompt}, which refers to the essay-writing instructions, from an \textit{input prompt}, which is the formatted text structure used to elicit responses from the language model. 

Let $\mathcal{T}$ denote the complete set of traits in the dataset, and let $\mathcal{V}_q \subseteq \mathcal{T}$ denote the traits with valid scores for prompt $q$. For each valid trait $j \in \mathcal{V}_q$, the gold score is $s_j$ with range $[l_{q,j},u_{q,j}]$.

We formulate multi-trait AES as autoregressive structured generation. Given a fixed trait order, the model generates a sequence of trait-value pairs:
\begin{equation}
o = [\,\mathrm{Trait}_j\!:\!v_j\,]_{j=1}^{|\mathcal{T}|}
\end{equation}
For traits that are valid for prompt $q$, $v_j$ is the predicted score $\hat{s}_j$; for traits that are not applicable to the prompt, $v_j$ is marked as \texttt{NaN}. The output is considered valid only if all traits appear in the predefined order, valid traits receive scores within their corresponding ranges, and invalid traits are marked as \texttt{NaN}. The training objective is to make the predicted scores $\hat{\mathbf{s}}_q=\{\hat{s}_j\}_{j\in\mathcal{V}_q}$ match the gold scores $\mathbf{s}_q=\{s_j\}_{j\in\mathcal{V}_q}$.

\subsection{GRPO-Based Post-Training}

Given the same input $z_q$, GRPO samples a group of $G$ candidate outputs:
\begin{equation}
\mathcal{O}=\{o_1,\ldots,o_G\}, \quad o_i\sim\pi_\theta(\cdot\mid z_q)
\end{equation}
In standard GRPO, each output receives a scalar reward $R_i$, and its group-normalized advantage is
\begin{equation}
A_i=\mathrm{Norm}_G(R_i)
=
\frac{R_i-\mu(\{R_k\}_{k=1}^G)}
{\sigma(\{R_k\}_{k=1}^G)+\epsilon}
\end{equation}
This scalar advantage is shared by all tokens in $o_i$. However, in multi-trait AES, a response may be correct on some traits but incorrect on others. A single sequence-level advantage cannot indicate which trait prediction should be reinforced or penalized. TAPO keeps the group-relative optimization framework but replaces the shared sequence-level advantage with token-level advantages that incorporate trait-specific feedback.

\section{TAPO}
\label{sec:method}

\begin{figure*}[t]
\centering
\includegraphics[width=\textwidth]{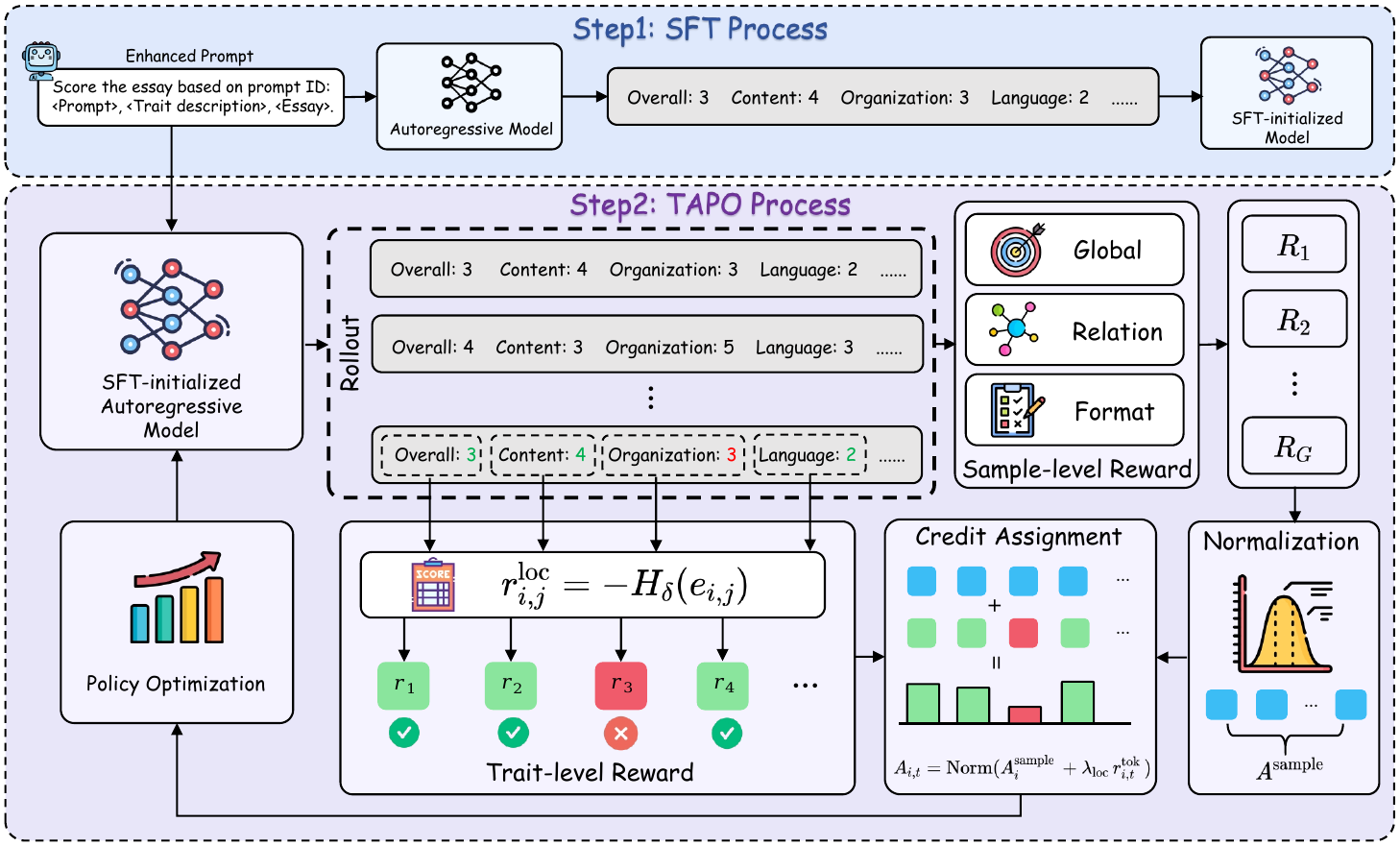}
\caption{
Overview of TAPO. 
After SFT with enhanced prompts, TAPO samples multiple structured trait-score sequences and computes both sample-level rewards and trait-level rewards. 
The normalized sample-level advantage is combined with trait-level credit assignment to provide fine-grained optimization signals for autoregressive multi-trait score generation.
}
\label{figure2}
\end{figure*}

TAPO is a trait-aware post-training framework for autoregressive multi-trait essay scoring. As shown in Figure~\ref{figure2}, TAPO integrates enhanced prompts, multi-dimensional reward design, and trait-level credit assignment to provide localized advantages for structured score generation. In particular, our reward design incorporates both score-matching errors and relative-order information across traits with respect to the ground-truth multi-trait score vector, enabling more fine-grained and trait-sensitive learning signals through localized supervision during post-training.
% it combines rubric-enhanced prompting, prompt-specific LoRA routing, multi-dimensional reward design, and trait-level credit assignment. 
% The reward design provides both sample-level guidance for global score-vector quality and trait-level feedback for individual scoring dimensions, enabling TAPO to optimize structured score generation with more localized learning signals.

\subsection{LLM-Enhanced Prompt}
\label{sec:trait_prompting}

During training, we enhance the model input with richer semantic information. Instead of using only the prompt ID and trait names \citep{doAutoregressiveScoreGeneration2024}, we include the original prompt text and append LLM-generated natural-language descriptions for the scoring traits. The model is trained to generate the structured trait-value sequence defined in Section~\ref{sec:problem_formulation}, using gold scores for traits applicable to the prompt and \texttt{NaN} for inapplicable traits.

This enhanced input provides both task-level context from the original prompt and trait-level semantics from the generated descriptions, enabling the model to better understand what each output position is expected to evaluate during structured multi-trait score generation.

\subsection{Reward Design}

TAPO adopts a sample-level reward to evaluate the predicted score vectors:
\begin{equation}
R_i^{\mathrm{sample}}
=
R_i^{\mathrm{global}}
+
\lambda_1 R_i^{\mathrm{rel}}
+
\lambda_2 R_i^{\mathrm{fmt}}
\end{equation}

\paragraph{Global reward.}
The global reward measures the overall agreement between the predicted and gold score vectors. Let $e_{i,j}$ denote the prediction error of trait $j$ normalized by its score range. We define
\begin{equation}
R_i^{\mathrm{global}}
=
-\frac{1}{|\mathcal{V}_q|}
\sum_{j\in\mathcal{V}_q}
\left[
\alpha H_\delta(e_{i,j})
+
\beta |e_{i,j}|
\right]
\end{equation}
where $\mathcal{V}_q$ is the set of valid traits for prompt $q$, and $H_\delta(\cdot)$ denotes the standard Huber penalty. The Huber penalty behaves quadratically for small errors and linearly for errors larger than the threshold $\delta$, with details provided in Appendix~\ref{app:huber_penalty}. Range normalization prevents traits with larger score ranges from dominating the reward computation, while the Huber-MAE combination provides smooth optimization near small errors and remains robust to larger deviations in predicted trait scores.

\paragraph{Relation reward.}
Motivated by pairwise ranking objectives \citep{burgesLearningRankUsing2005}, we use a rank-consistency reward to capture inter-trait consistency. After normalizing trait scores to $[0,1]$, the relative order reflects the essay's trait-level quality profile rather than the intrinsic importance of traits. Thus, if a trait is clearly higher than another in the gold scores but the prediction reverses this pattern, we regard it as a severe trait-level inconsistency.

Let $\tilde{s}_j$ and $\tilde{\hat{s}}_{i,j}$ denote normalized gold and predicted scores. We consider trait pairs whose gold scores are different:
\begin{equation}
\mathcal{P}_q
=
\{(a,b): a,b\in\mathcal{V}_q,\; a\prec b,\; \tilde{s}_a \neq \tilde{s}_b\}
\end{equation}
where $\prec$ denotes the predefined trait order used only to avoid duplicate pairs. For each pair $(a,b)$, let $m_{ab}$ be the gold ranking direction and $\Delta_{i,ab}$ be the predicted score gap. The relation reward is
\begin{equation}
R_i^{\mathrm{rel}}
=
-\frac{1}{|\mathcal{P}_q|}
\sum_{(a,b)\in\mathcal{P}_q}
\max(0,-m_{ab}\Delta_{i,ab})
\end{equation}
where $m_{ab}=\mathrm{sign}(\tilde{s}_a-\tilde{s}_b)$ and $\Delta_{i,ab}=\tilde{\hat{s}}_{i,a}-\tilde{\hat{s}}_{i,b}$. If $\mathcal{P}_q$ is empty, we set $R_i^{\mathrm{rel}}=0$.

\paragraph{Format reward.}
The format term $R_i^{\mathrm{fmt}}$ penalizes malformed structured outputs that cannot be reliably parsed. We set $R_i^{\mathrm{fmt}}=0$ when the output follows the predefined format, and $R_i^{\mathrm{fmt}}=-1$ otherwise. A valid output must contain all traits in the predefined order, assign exactly one value to each trait, use scores within the valid range for applicable traits, and mark inapplicable traits as \texttt{NaN}, ensuring consistent structured score generation.

\begin{table}[t]
\centering
\scriptsize
\setlength{\tabcolsep}{2.6pt}
\renewcommand{\arraystretch}{1.05}
\begin{adjustbox}{max width=\columnwidth}
\begin{tabular}{c|c|l|r|c}
\toprule
\textbf{Dataset} & \textbf{Pr} & \multicolumn{1}{c|}{\textbf{Traits}} & \multicolumn{1}{c|}{\textbf{Es}} & \textbf{Range} \\
\midrule
\multirow{8}{*}{\rotatebox{90}{ASAP/ASAP++}}
& 1 & Over, Cont, Org, WC, SF, Conv        & 1,783 & 2--12 / 1--6  \\
& 2 & Over, Cont, Org, WC, SF, Conv        & 1,800 & 1--6 / 1--6   \\
& 3 & Over, Cont, PA, Nar, Lang            & 1,726 & 0--3 / 0--3   \\
& 4 & Over, Cont, PA, Nar, Lang            & 1,772 & 0--3 / 0--3   \\
& 5 & Over, Cont, PA, Nar, Lang            & 1,805 & 0--4 / 0--4   \\
& 6 & Over, Cont, PA, Nar, Lang            & 1,800 & 0--4 / 0--4   \\
& 7 & Over, Cont, Org, Conv, Style         & 1,569 & 0--30 / 0--6  \\
& 8 & Over, Cont, Org, WC, SF, Conv, Voice & 723   & 0--60 / 2--12 \\
\midrule
Feedback
& -- & Coh, Syn, Voc, Phr, Gram, Conv       & 3,930 & -- / 1--5     \\
\bottomrule
\end{tabular}
\end{adjustbox}
\caption{
Dataset statistics for ASAP/ASAP++ and Feedback Prize.
Pr: prompt number; Es: number of essays; Range: Overall / Trait score range.
Over: \textit{Overall}; Cont: \textit{Content}; Org: \textit{Organization}; WC: \textit{Word Choice}; SF: \textit{Sentence Fluency}; Conv: \textit{Conventions}; PA: \textit{Prompt Adherence}; Nar: \textit{Narrativity}; Lang: \textit{Language};
Coh: \textit{Cohesion}; Syn: \textit{Syntax}; Voc: \textit{Vocabulary}; Phr: \textit{Phraseology}; Gram: \textit{Grammar}.
}
\label{tab:datasets}
\end{table}

\subsection{Trait-Level Credit Assignment}
\label{sec:trait_credit}

The sample-level reward evaluates the overall score vector, but it cannot identify which trait is responsible for an error. To provide localized feedback, we define a local reward for each valid trait:
\begin{equation}
r_{i,j}^{\mathrm{loc}} = -H_\delta(e_{i,j}),
\label{eq:local_reward}
\end{equation}
where $j \in \mathcal{V}_q$ and $H_\delta(\cdot)$ denotes the Huber penalty defined in Appendix~\ref{app:huber_penalty}.

For each input, we normalize the local rewards across the $G$ sampled outputs for the same trait. This makes trait-level rewards comparable within the group and prevents traits with different error scales from dominating the local update. The normalized local reward for trait $j$ is then assigned to the ending token of its generated score span. Non-score tokens receive no additional local reward.

Let $A_i^{\mathrm{sample}}$ denote the group-normalized sample-level advantage derived from $R_i^{\mathrm{sample}}$, and let $r_{i,t}^{\mathrm{tok}}$ denote the normalized local reward injected at token $t$. TAPO constructs the final token-level advantage as
\begin{equation}
A_{i,t}
=
\mathrm{Norm}
\left(
A_i^{\mathrm{sample}}
+
\lambda_{\mathrm{loc}} r_{i,t}^{\mathrm{tok}}
\right)
\label{eq:token_advantage}
\end{equation}
where $\mathrm{Norm}$ denotes normalization over the sampled group for the same input. The resulting advantages are used in the GRPO policy objective. Compared with standard GRPO, which assigns the same advantage to all tokens in a response, TAPO preserves global sample-level guidance while adding localized feedback to trait-specific score tokens.

\begin{table*}[t]
\centering
\scriptsize
\setlength{\tabcolsep}{4pt}
\renewcommand{\arraystretch}{1.08}
\begin{adjustbox}{max width=\textwidth}
\begin{tabular}{l|ccccccccccc|c}
\toprule
& \multicolumn{11}{c|}{\textbf{Traits (Prediction Order: $\leftarrow$)}} & \\
\midrule
\textbf{Method} 
& \textbf{Overall} 
& \textbf{Content} 
& \textbf{PA} 
& \textbf{Lang} 
& \textbf{Nar} 
& \textbf{Org} 
& \textbf{Conv} 
& \textbf{WC} 
& \textbf{SF} 
& \textbf{Style} 
& \textbf{Voice} 
& \textbf{Avg.} \\
\midrule
T5-large + SFT  
& 0.754 & 0.743 & 0.747 & 0.696 & 0.721 & 0.702 & 0.695 & 0.700 & 0.698 & 0.727 & 0.612 & 0.709 \\
+GRPO 
& 0.776 & 0.754 & 0.756 & 0.710 & 0.726 & \textbf{0.720} & 0.718 & 0.700 & 0.705 & \textbf{0.732} & 0.597 
& 0.718 {\color{green1}{(+0.009)}} \\
\rowcolor{blue1}
+TAPO 
& \textbf{0.777} & \textbf{0.760} & \textbf{0.764} & \textbf{0.712} & \textbf{0.734} & 0.714 & \textbf{0.728} & \textbf{0.725} & \textbf{0.711} & 0.717 & \textbf{0.642} 
& \textbf{0.726} {\color{green1}{(+0.017)}} \\
\midrule
Qwen3-1.7B + SFT  
& 0.762 & 0.728 & 0.747 & 0.694 & \textbf{0.725} & 0.689 & 0.664 & 0.688 & 0.664 & 0.710 & \textbf{0.620} & 0.699 \\
+GRPO
& 0.761 & 0.725 & 0.746 & 0.691 & \textbf{0.725} & 0.685 & 0.660 & 0.685 & 0.661 & 0.707 & 0.608 
& 0.696 {\color{gray}{(-0.003)}} \\
\rowcolor{blue1}
+TAPO 
& \textbf{0.770} & \textbf{0.736} & \textbf{0.753} & \textbf{0.701} & 0.722 & \textbf{0.699} & \textbf{0.685} & \textbf{0.698} & \textbf{0.683} & \textbf{0.733} & 0.600 
& \textbf{0.707} {\color{green1}{(+0.008)}} \\
\bottomrule
\end{tabular}
\end{adjustbox}
\caption{
Trait-wise comparison of SFT, GRPO, and TAPO on ASAP and ASAP++ datasets under five-fold cross-validation. Scores are averaged over prompts for each trait. 
Traits are predicted from right to left ($\leftarrow$). }
\label{tab:post_training_effectiveness}
\end{table*}

\begin{table*}[t]
\centering
\scriptsize
\setlength{\tabcolsep}{4pt}
\renewcommand{\arraystretch}{1.05}
\begin{adjustbox}{max width=\textwidth}
\begin{tabular}{l|ccccccccccc|c}
\toprule
& \multicolumn{11}{c|}{\textbf{Traits (Prediction Order: $\leftarrow$)}} & \\
\midrule
\textbf{Method} 
& \textbf{Overall} 
& \textbf{Content} 
& \textbf{PA} 
& \textbf{Lang} 
& \textbf{Nar} 
& \textbf{Org} 
& \textbf{Conv} 
& \textbf{WC} 
& \textbf{SF} 
& \textbf{Style} 
& \textbf{Voice} 
& \textbf{Avg.} \\
\midrule
HISK        
& 0.718 & 0.679 & 0.697 & 0.605 & 0.659 & 0.610 & 0.527 & 0.579 & 0.553 & 0.609 & 0.489 & 0.611 \\
MTL-BiLSTM  
& 0.764 & 0.685 & 0.701 & 0.604 & 0.668 & 0.615 & 0.560 & 0.615 & 0.598 & 0.632 & 0.582 & 0.638 \\
STL-LSTM    
& 0.750 & 0.707 & 0.731 & 0.640 & 0.699 & 0.649 & 0.605 & 0.621 & 0.612 & 0.659 & 0.544 & 0.656 \\
DualTrans   
& \textbf{0.778} & 0.726 & 0.732 & 0.660 & 0.704 & 0.682 & 0.668 & 0.674 & 0.663 & 0.689 & 0.619 & 0.687 \\
T-MES       
& 0.774 & 0.730 & 0.750 & 0.702 & 0.730 & 0.685 & 0.686 & 0.679 & 0.675 & 0.693 & 0.590 & 0.700 \\
ArTS        
& 0.752 & 0.729 & 0.749 & 0.701 & 0.727 & 0.677 & 0.683 & 0.683 & 0.683 & 0.712 & 0.621 & 0.702 \\
SaMRL       
& 0.754 & 0.735 & 0.751 & 0.703 & 0.728 & 0.682 & 0.685 & 0.688 & 0.691 & 0.710 & 0.627 & 0.705 \\
RMTS        
& 0.755 & 0.737 & 0.752 & \textbf{0.713} & \textbf{0.744} & 0.682 & 0.690 & 0.705 & \underline{0.694} & 0.702 & 0.612 & 0.708 \\
GAT-AES    
& 0.771 & 0.742 & 0.749 & 0.687 & 0.726 & 0.694 & 0.686 & \underline{0.709} & 0.692 & 0.699 & \textbf{0.649} & 0.710 \\
RaDME      
& 0.754 & \underline{0.744} & \underline{0.759} & 0.706 & \underline{0.736} & \underline{0.701} & \underline{0.692} & 0.693 & 0.692 & \textbf{0.719} & 0.623 & \underline{0.711} \\
\midrule
TAPO        
& \underline{0.777} & \textbf{0.760} & \textbf{0.764} & \underline{0.712} & 0.734 & \textbf{0.714} & \textbf{0.728} & \textbf{0.725} & \textbf{0.711} & \underline{0.717} & \underline{0.642} & \textbf{0.726} \\
\bottomrule
\end{tabular}
\end{adjustbox}
\caption{
Trait-wise QWK results on ASAP and ASAP++. Each value is averaged over the prompts where the corresponding trait is annotated. The best and second-best results among reported methods are shown in bold and underlined, respectively.}
\label{tab:trait_main_results}
\end{table*}

\begin{table}[t]
\centering
\scriptsize
\setlength{\tabcolsep}{3.2pt}
\renewcommand{\arraystretch}{1.05}
\begin{adjustbox}{max width=\columnwidth}
\begin{tabular}{l|c|c|c|c|c|c|c}
\toprule
& \multicolumn{6}{c|}{\textbf{Traits (Prediction Order: $\leftarrow$)}} & \\
\midrule
\textbf{Model} 
& \textbf{Coh} 
& \textbf{Syn} 
& \textbf{Voc} 
& \textbf{Phr} 
& \textbf{Gram} 
& \textbf{Conv} 
& \textbf{AVG} \\
\midrule
MTL$^{*}$   
& 0.462 & 0.507 & 0.519 & 0.505 & 0.484 & 0.527 & 0.501 \\
ArTS$^{*}$  
& 0.590 & 0.628 & 0.594 & 0.639 & 0.659 & 0.659 & 0.628 \\
SaMRL$^{*}$ 
& 0.592 & 0.637 & \textbf{0.610} & 0.646 & 0.658 & 0.656 & 0.633 \\
\midrule
TAPO
& \textbf{0.603} & \textbf{0.656} & 0.609 & \textbf{0.651} & \textbf{0.664} & \textbf{0.682} & \textbf{0.644} \\
\bottomrule
\end{tabular}
\end{adjustbox}
\caption{
Results on the Feedback Prize dataset. Five-fold averaged QWK score is reported.
Coh: \textit{Cohesion}; Syn: \textit{Syntax}; Voc: \textit{Vocabulary}; Phr: \textit{Phraseology}; Gram: \textit{Grammar}; Conv: \textit{Conventions}.
Results marked with $^{*}$ are reported by \citet{doAutoregressiveMultitraitEssay2024}.
}
\label{tab:feedback_prize}
\end{table}

\begin{table*}[t]
\centering
\scriptsize
\setlength{\tabcolsep}{4pt}
\renewcommand{\arraystretch}{1.05}
\begin{adjustbox}{max width=\textwidth}
\begin{tabular}{l|ccccccccccc|c}
\toprule
& \multicolumn{11}{c|}{\textbf{Traits (Prediction Order: $\leftarrow$)}} & \\
\midrule
\textbf{Setting} 
& \textbf{Overall} 
& \textbf{Content} 
& \textbf{PA} 
& \textbf{Lang} 
& \textbf{Nar} 
& \textbf{Org} 
& \textbf{Conv} 
& \textbf{WC} 
& \textbf{SF} 
& \textbf{Style} 
& \textbf{Voice} 
& \textbf{AVG} \\
\midrule
Full model 
& 0.777 
& \textbf{0.760} 
& \textbf{0.764} 
& \textbf{0.712} 
& 0.734 
& 0.714 
& \textbf{0.728} 
& \textbf{0.725} 
& 0.711 
& 0.717 
& 0.642 
& \textbf{0.726} \\
\midrule
w/o trait reward
& \textbf{0.784} 
& 0.746 
& 0.743 
& 0.686 
& 0.730 
& \textbf{0.722} 
& 0.721 
& 0.715 
& 0.716 
& 0.712 
& 0.656 
& 0.721 \\
w/o $R_{\mathrm{relation}}$
& 0.779 
& 0.757 
& 0.754 
& 0.695 
& 0.729 
& 0.720 
& 0.725 
& 0.719 
& \textbf{0.718} 
& 0.717 
& 0.649 
& 0.724 \\
w/o $R_{\mathrm{format}}$
& 0.775 
& 0.757 
& 0.753 
& 0.697 
& \textbf{0.739} 
& 0.717 
& 0.717 
& 0.713 
& 0.706 
& 0.721 
& \textbf{0.658} 
& 0.723 \\
w/o enhanced prompt
& 0.764
& 0.749
& 0.755
& 0.690
& 0.727
& 0.717
& 0.703
& 0.707
& 0.701
& \textbf{0.732}
& 0.631
& 0.716 \\
\bottomrule
\end{tabular}
\end{adjustbox}
\caption{
Trait-wise ablation results on ASAP and ASAP++. Scores are averaged over prompts for each trait. Full model denotes the complete version of TAPO, and the best result in each column is shown in bold.
}
\label{tab:trait_ablation}
\end{table*}

\section{Experimental Setup}
\label{sec:experiments}

\paragraph{Datasets.}
We use ASAP\footnote{\url{https://www.kaggle.com/c/asap-aes}} and ASAP++\footnote{\url{https://lwsam.github.io/ASAP++/lrec2018.html}} as our main benchmarks. ASAP contains essays from eight prompts with holistic scores, while ASAP++ extends it with fine-grained trait annotations for multi-trait essay scoring \citep{mathiasASAPEnrichingASAP2018}. The applicable traits and score ranges vary across prompts. We also evaluate TAPO on Feedback Prize\footnote{\url{https://www.kaggle.com/competitions/feedback-prize-english-language-learning}}, which contains argumentative essays annotated with six writing traits. Following prior multi-trait AES work, we use 5-fold cross-validation for all experiments \citep{doAutoregressiveScoreGeneration2024,doAutoregressiveMultitraitEssay2024}. Dataset statistics and trait configurations are shown in Table~\ref{tab:datasets}.

\paragraph{Backbone models.}
We use T5-large as the main backbone \citep{raffelExploringLimitsTransfer2020}, following prior autoregressive multi-trait scoring work \citep{doAutoregressiveScoreGeneration2024,doAutoregressiveMultitraitEssay2024}. To examine whether TAPO remains effective on a decoder-only backbone, we further evaluate TAPO with Qwen3-1.7B, a decoder-only causal language model from the Qwen family \citep{yangQwen3TechnicalReport2025}.

\paragraph{Evaluation metric.}
We use Quadratic Weighted Kappa (QWK), the standard metric for AES, which measures agreement with human scores while penalizing larger score discrepancies more heavily. We report trait-wise and prompt-wise QWK. Trait-wise results are computed over valid prompt-trait pairs for each trait, while prompt-wise results average over applicable traits within each prompt. Final scores are macro-averaged over traits or prompts.

\paragraph{Baselines.}
We compare TAPO with representative multi-trait AES methods, including feature-based and recurrent models (HISK, STL-LSTM, MTL-BiLSTM) \citep{cozmaAutomatedEssayScoring2018,mathiasCanNeuralNetworks2020}, Transformer-based and structured trait-relation models (DualTrans, T-MES, GAT-AES) \citep{DualscaleBERTUsing2024,wangTMESTraitAwareMixofExperts2025,liGraphBasedMultiTraitEssay2025}, and recent autoregressive, RL-based, and LLM-enhanced methods (ArTS, SaMRL, RMTS, RaDME) \citep{doAutoregressiveScoreGeneration2024,doAutoregressiveMultitraitEssay2024,chuRationaleEssayScores2025,doTeachtoreasonScoringSelfexplainable2026}. For the standard GRPO baseline, we use a single MAE-based sequence-level score-agreement reward over the generated score vector.

\paragraph{Implementation details.}
During training, we append trait descriptions generated by Gemini-3-Pro. For T5-large, we train for 20 epochs with batch size 16 and learning rate $3\times10^{-5}$, and use the best validation checkpoint to initialize post-training. TAPO is instantiated with GRPO, using group size $G=4$, 2 training epochs, and prompt-specific LoRA adapters with rank $r=32$. Reward hyperparameters are selected on the validation set, with sensitivity analyses in Section~\ref{sec:hyperparameter_analysis}. Experiments are conducted on NVIDIA A100 GPUs. Qwen3-1.7B details are provided in Appendix~\ref{app:qwen_implementation}.

\section{Results and Analysis}

\subsection{Effectiveness of TAPO}
\label{sec:post_training_effectiveness}

We conduct experiments on the ASAP and ASAP++ datasets using T5-large and Qwen3-1.7B. This comparison directly examines whether the proposed trait-aware post-training strategy improves over standard GRPO under the same autoregressive multi-trait scoring setting. Table~\ref{tab:post_training_effectiveness} reports the trait-wise results, and the corresponding prompt-wise results are provided in Appendix~\ref{app:prompt_results}. TAPO achieves the best average trait-wise QWK under both backbones. On T5-large, GRPO improves the average QWK from 0.709 to 0.718, while TAPO further raises it to 0.726. On Qwen3-1.7B, GRPO slightly underperforms SFT, decreasing the average from 0.699 to 0.696, whereas TAPO improves it to 0.707 and achieves the best results on most traits. These results indicate that standard sequence-level GRPO is not uniformly
beneficial for structured multi-trait generation across backbones, while
the proposed trait-aware reward design and credit assignment provide more
effective post-training signals.

Since GRPO involves stochastic rollout sampling, we additionally conduct
three-seed comparisons between TAPO and standard GRPO under identical enhanced prompts. TAPO improves Mean QWK from 0.7272 to 0.7316, with consistent gains across prompt-trait, trait, and prompt-level comparisons. Detailed results and paired significance analyses are provided in Appendix~\ref{sec:multiseed}.

\subsection{Overall Performance}
\label{sec:performance_comparison}

%\subsubsection{Performance on ASAP and ASAP++}
\paragraph{Performance on ASAP and ASAP++.}
\label{sec:asap_comparison}

Table~\ref{tab:trait_main_results} reports the trait-wise QWK results on
ASAP and ASAP++. The corresponding prompt-wise results are provided in
Appendix~\ref{app:prompt_results}. TAPO achieves the best average performance,
with an average QWK of 0.726. Compared with the strongest previously reported
result, RaDME, TAPO improves the average QWK by 0.015. It achieves the best
result on 6 out of 11 traits and ranks second on four additional traits, with
Narrativity being the only trait where it does not rank among the top two.
This pattern indicates that the improvements are broadly distributed across
traits rather than concentrated on a small subset. Additional paired analyses
against RaDME, including a comparison without enhanced prompts, are provided
in Appendix~\ref{sec:radme_analysis}.

Compared with autoregressive and RL-based baselines, TAPO also shows consistent
gains. It improves over ArTS by 0.024 average QWK and over SaMRL by 0.021,
suggesting that trait-aware post-training provides benefits beyond
autoregressive score generation and sequence-level RL rewards. Standard
deviation comparisons are reported in Appendix~\ref{app:sd_results}.

%\subsubsection{Performance on Feedback Prize}
\paragraph{Performance on Feedback Prize.}
\label{sec:feedback_prize}

We further evaluate TAPO on Feedback Prize. This experiment examines whether TAPO remains effective on a dataset with a different trait schema and data distribution.

As shown in Table~\ref{tab:feedback_prize}, TAPO achieves the best average QWK compared with the reported baselines on Feedback Prize, outperforming SaMRL by 0.011 average QWK. It obtains the best performance on 5 out of 6 traits, with particularly clear gains on Cohesion, Syntax, Phraseology, and Conventions. These results suggest that TAPO is not limited to ASAP and ASAP++, but can also be applied effectively to another multi-trait essay scoring dataset with distinct annotation criteria.

\subsection{Ablation Study}
\label{sec:ablation}

Table~\ref{tab:trait_ablation} reports the trait-wise ablation results, with the corresponding prompt-wise results provided in Appendix~\ref{app:prompt_results}. Removing the trait-level reward reduces the average QWK from 0.726 to 0.721, indicating that localized trait feedback provides useful supervision beyond sample-level rewards. However, the effect is not monotonic across individual traits: several traits, including Overall, perform better in this ablation. We therefore view localized trait feedback as a complementary component of TAPO rather than the sole source of improvement. Since TAPO jointly optimizes the complete structured output, improvements in aggregate scoring quality and inter-trait consistency may involve local trade-offs among individual traits.

The relation and format rewards also contribute to performance. Without $R_{\mathrm{relation}}$, the average QWK drops to 0.724, suggesting that inter-trait consistency complements direct score matching. Removing $R_{\mathrm{format}}$ yields an average QWK of 0.723, showing that format validity acts as a stabilizing constraint for structured generation.

Removing the enhanced prompt causes the largest decrease, reducing the average QWK to 0.716. This suggests that enhancing the input with original prompt texts and LLM-generated trait descriptions helps the model better capture prompt and trait semantics. Overall, although some ablated variants perform better on individual traits, the full model achieves the best average performance.

\paragraph{Prompt-source analysis.}
Since removing the enhanced prompt causes the largest average performance
drop, we further examine the respective contributions of the original writing prompt and trait descriptions, as well as the effect of description source. As shown in Table~\ref{tab:prompt_source}, the prompt-only setting retains the original writing prompt but removes natural-language trait descriptions. The LLM-generated descriptions used in TAPO are produced by Gemini-3-Pro. For the human rubric-style setting, we summarize the released ASAP++ prompt-specific scoring guidelines for traits covered by the guidelines and use concise author-written rubric-style descriptions for uncovered traits.

Prompt-only input performs worse than both description-based settings,
indicating that trait descriptions provide useful semantic guidance beyond
the original writing prompt. Human rubric-style descriptions recover part of the gap, showing that the benefit is not exclusive to LLM-generated
descriptions. LLM-generated descriptions achieve the best performance in our experiments, indicating some sensitivity to the source and construction of trait descriptions.

\begin{table}[t]
\centering
\scriptsize
\setlength{\tabcolsep}{3.2pt}
\renewcommand{\arraystretch}{1.05}
\begin{adjustbox}{max width=\columnwidth}
\begin{tabular}{l|ccc}
\toprule
\textbf{Input setting}
& \textbf{Prompt Avg.}
& \textbf{Trait Avg.}
& \textbf{Mean} \\
\midrule
Prompt only
& 0.7355
& 0.7198
& 0.7277 \\
Human descriptions
& 0.7390
& 0.7226
& 0.7308 \\
LLM-generated descriptions
& \textbf{0.7427}
& \textbf{0.7257}
& \textbf{0.7342} \\
\bottomrule
\end{tabular}
\end{adjustbox}
\caption{
Effect of different enhanced-input settings on ASAP and ASAP++.
Prompt Avg. and Trait Avg. denote prompt-wise and trait-wise QWK, respectively.
}
\label{tab:prompt_source}
\end{table}

\subsection{Hyperparameter Analysis}
\label{sec:hyperparameter_analysis}

\begin{figure}[t]
\centering
\includegraphics[width=\columnwidth]{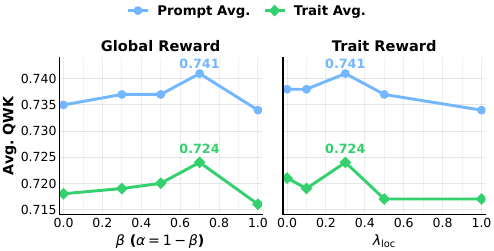}
\caption{
Hyperparameter analysis of TAPO. 
The left panel varies $\alpha$ and $\beta$ in the global reward, while the right panel varies the local trait-level reward weight $\lambda_{\mathrm{loc}}$.
}
\label{fig:hyperparameter_analysis}
\end{figure}

Figure~\ref{fig:hyperparameter_analysis} analyzes two key hyperparameters in TAPO. These experiments are conducted without $R_{\mathrm{relation}}$ to isolate the effects of the global reward formulation and the local trait-level reward weight. In the left panel, using only Huber or only MAE yields lower Trait Avg. scores of 0.718 and 0.716, respectively. Combining the two terms performs better, with the best result achieved at $\alpha=0.3$ and $\beta=0.7$ (0.741 Prompt Avg. and 0.724 Trait Avg.), suggesting that an MAE-dominant reward provides robust global guidance while the Huber term smooths optimization.

The right panel shows that a moderate local reward weight works best. Removing the trait reward or using a very small weight underperforms, while larger weights such as 0.5 and 1.0 also degrade performance. This indicates that trait-level feedback is useful but should be balanced with sample-level rewards. Additional analysis of the relation reward weight is provided in Appendix~\ref{app:additional_hyperparameter}; the format reward weight is fixed to 0.1 as a small auxiliary constraint for structured output validity.

\subsection{Qualitative Analysis} \label{sec:qualitative_analysis} 
To illustrate how TAPO differs from standard GRPO, we visualize the token-level credit assigned to generated trait-score sequences. In GRPO, the same sequence-level advantage is applied to all output tokens within a sampled response. Therefore, even if only a subset of trait scores is incorrect, all trait predictions receive an identical optimization signal. In contrast, TAPO combines the sample-level advantage with trait-level local feedback, allowing erroneous trait-score tokens to receive lower effective credit. 

\begin{figure}[t]
\centering
\includegraphics[width=\columnwidth]{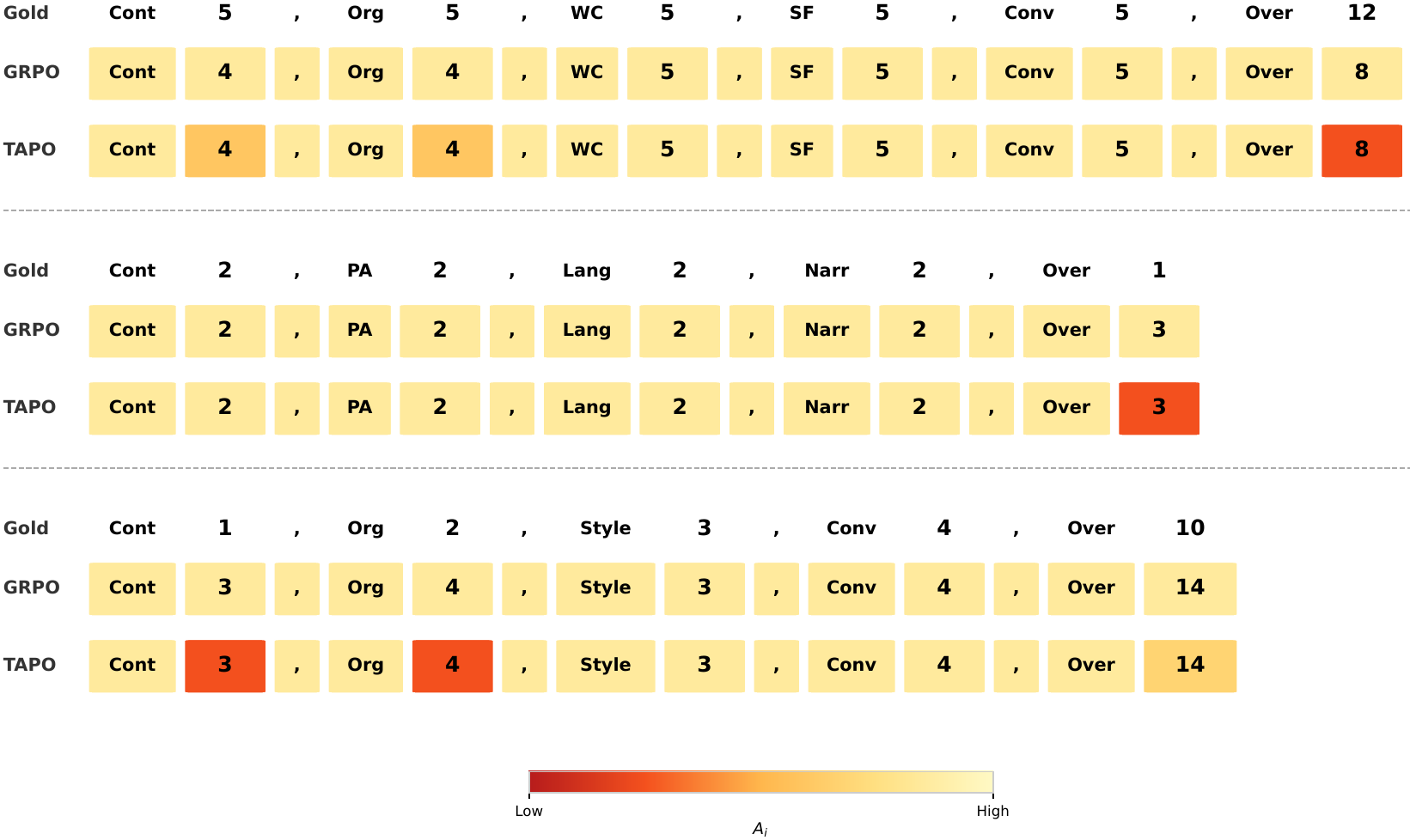}
\caption{
Qualitative visualization of token-level credit assignment.
The three blocks, from top to bottom, correspond to examples from Prompt 1, Prompt 3, and Prompt 7.
GRPO and TAPO generate the same score sequence in each block, but assign different token-level credits.
Darker colors indicate lower effective credit, while lighter colors indicate higher effective credit.
}
\label{fig:credit_assignment}
\end{figure}

Figure~\ref{fig:credit_assignment} shows three representative examples from different prompts and trait sets. In each example, the GRPO and TAPO rows contain the same generated score sequence, while the background color indicates the assigned credit. GRPO produces nearly uniform coloring across all score tokens, reflecting its sequence-level credit assignment. TAPO preserves higher credit for correctly predicted traits and reduces the credit of erroneous traits. For example, when the overall score has a larger error, its corresponding score token receives a darker color, indicating stronger negative local feedback. These examples show that TAPO provides more localized optimization signals for structured multi-trait score generation.

\subsection{Comparison with Per-Trait Training}
\label{sec:per_trait}

To examine how joint structured generation compares with trait-specific
prediction, we construct per-trait SFT and per-trait MTL+RL baselines.
Each essay is decomposed into trait-specific instances, such that the model
generates one trait score at a time. For per-trait MTL+RL, each GRPO group
contains multiple rollouts for the same essay-trait pair and is optimized
using a trait-specific sample-level reward. Since each output contains only
one trait prediction, no trait-level token credit assignment is required.
We further consider an inter-trait consistency reward, weighted by
$\lambda_{\mathrm{cons}}$. Specifically, the independently generated trait
scores for the same essay are first normalized by their corresponding score
ranges, and a pairwise hinge reward is computed according to the gold
inter-trait ordering. This consistency reward is then added to the
trait-specific reward, encouraging independently generated scores to preserve
the relative trait profile of the essay.

As shown in Table~\ref{tab:per_trait}, per-trait SFT already provides a strong
baseline. Applying RL yields modest further improvements, while introducing
the consistency reward increases mean QWK from 0.7303 to 0.7307, providing a
small additional gain from modeling inter-trait dependencies. TAPO achieves
the highest point estimate, with a mean QWK of 0.7342. However, its differences
from the per-trait MTL+RL baselines are not statistically significant under
five-fold paired tests. These results show that TAPO remains competitive with
explicit per-trait optimization while jointly generating the complete
multi-trait score sequence, but do not establish a statistically significant
advantage over the per-trait alternatives.

\begin{table}[t]
\centering
\scriptsize
\setlength{\tabcolsep}{2.8pt}
\renewcommand{\arraystretch}{1.05}
\begin{adjustbox}{max width=\columnwidth}
\begin{tabular}{l|ccc}
\toprule
\textbf{Method}
& \textbf{Prompt Avg.}
& \textbf{Trait Avg.}
& \textbf{Mean} \\
\midrule
Per-trait SFT
& 0.7373
& 0.7220
& 0.7297 \\
Per-trait MTL+RL ($\lambda_{\mathrm{cons}}=0$)
& 0.7374
& 0.7231
& 0.7303 \\
Per-trait MTL+RL ($\lambda_{\mathrm{cons}}=0.2$)
& 0.7380
& 0.7234
& 0.7307 \\
\midrule
TAPO
& \textbf{0.7427}
& \textbf{0.7257}
& \textbf{0.7342} \\
\bottomrule
\end{tabular}
\end{adjustbox}
\caption{
Comparison with per-trait supervised and policy-optimization baselines
on ASAP and ASAP++. Mean denotes the arithmetic mean of Prompt Avg. and
Trait Avg.
}
\label{tab:per_trait}
\end{table}

\section{Conclusion}
\label{sec:conclusion}

We presented TAPO, a trait-aware post-training framework for autoregressive
multi-trait essay scoring. TAPO adapts policy optimization to structured score
generation by combining multi-dimensional sample-level rewards with
trait-level credit assignment, while enhanced prompts provide explicit
task- and trait-level semantic guidance. Experiments on ASAP/ASAP++ and
Feedback Prize with T5-large and Qwen3-1.7B show that our approach consistently
improves over supervised fine-tuning and standard GRPO. Further analyses show
that enhanced prompting contributes substantially to the final performance,
while localized trait feedback provides a smaller but complementary benefit.
Comparisons with per-trait training further show that joint structured
generation with TAPO remains competitive with explicit trait-specific
optimization. Overall, these results highlight the value of combining
semantic input enhancement with trait-aware policy optimization for
autoregressive multi-trait essay scoring.

\section*{Limitations}
\label{sec:limitations}

Although TAPO shows consistent improvements, our experiments remain limited
in scale. We evaluate the framework with T5-large and Qwen3-1.7B, but do not
explore substantially larger models or a wider hyperparameter space due to
computational constraints. Moreover, the improvement over standard GRPO under
matched enhanced prompts is statistically reliable but numerically modest,
and trait-level credit assignment does not improve every individual trait
monotonically. Our prompt-source analysis compares Gemini-3-Pro-generated
descriptions with human rubric-style descriptions, but is limited to one LLM
and one human-description construction strategy. We do not systematically
study alternative LLMs, generation prompts, or other ways of constructing
trait descriptions.

TAPO currently uses a simple additive formulation to combine sample-level and
trait-level advantages. While this design is straightforward and effective,
more adaptive credit-assignment mechanisms remain unexplored. In addition,
localized credit is derived primarily from trait-specific score errors,
whereas relation and format signals are still applied at the sample level
rather than decomposed into localized feedback. Exploring more expressive
credit-assignment strategies, broader model scales, and evaluations beyond
English essay-scoring benchmarks remains important future work.

\section*{Ethical Statement}
\label{sec:ethics}

This work studies automated multi-trait essay scoring. Such systems can support educational assessment by improving scoring efficiency and providing trait-level signals, but they should not be used as the sole basis for high-stakes decisions. Automated scoring models may inherit biases from training data or scoring rubrics, and their predictions can be affected by demographic, linguistic, or topical factors not fully captured by benchmark evaluations. Therefore, we recommend that TAPO be used as an assistive tool under human oversight, especially in high-stakes educational contexts.

All experiments are conducted on publicly available essay scoring datasets. We do not collect new personal data. The datasets are used only for research purposes, and we report aggregate evaluation results rather than individual student-level information.

\paragraph{Acknowledgements}
This work is supported by the National Natural Science Foundation of China
(62677001), and the Fundamental Research Funds for the Central Universities, Peking University.

% Bibliography entries for the entire Anthology, followed by custom entries
%\bibliography{anthology,custom}
% Custom bibliography entries only
\FloatBarrier
\bibliography{main}

\appendix

\section{Huber Penalty}
\label{app:huber_penalty}

\begin{table*}[t]
\centering
\scriptsize
\setlength{\tabcolsep}{6pt}
\renewcommand{\arraystretch}{1.08}
\begin{adjustbox}{max width=\textwidth}
\begin{tabular}{l|cccccccc|c}
\toprule
& \multicolumn{8}{c|}{\textbf{Prompts}} & \\
\midrule
\textbf{Method} 
& \textbf{P1} 
& \textbf{P2} 
& \textbf{P3} 
& \textbf{P4} 
& \textbf{P5} 
& \textbf{P6} 
& \textbf{P7} 
& \textbf{P8} 
& \textbf{Avg.} \\
\midrule
T5-large + SFT  
& 0.721 & 0.708 & 0.700 & 0.776 & 0.715 & 0.763 & 0.757 & 0.661 & 0.725 \\
+GRPO
& \textbf{0.739} & \textbf{0.745} & 0.708 & \textbf{0.782} & \textbf{0.728} & 0.780 & 0.761 & 0.663
& 0.738 {\color{green1}{(+0.013)}} \\
\rowcolor{blue1}
+TAPO
& 0.730 & 0.737 & \textbf{0.713} & \textbf{0.782} & \textbf{0.728} & \textbf{0.793} & \textbf{0.765} & \textbf{0.694}
& \textbf{0.743} {\color{green1}{(+0.018)}} \\
\midrule
Qwen3-1.7B + SFT  
& 0.717 & 0.688 & \textbf{0.717} & 0.758 & 0.732 & 0.763 & 0.728 & 0.638 & 0.717 \\
+GRPO
& 0.714 & 0.685 & 0.716 & 0.758 & 0.731 & 0.762 & 0.725 & 0.632
& 0.715 {\color{gray}{(-0.002)}} \\
\rowcolor{blue1}
+TAPO
& \textbf{0.721} & \textbf{0.711} & 0.715 & \textbf{0.759} & \textbf{0.737} & \textbf{0.766} & \textbf{0.747} & \textbf{0.651}
& \textbf{0.726} {\color{green1}{(+0.009)}} \\
\bottomrule
\end{tabular}
\end{adjustbox}
\caption{
Prompt-wise comparison of SFT, GRPO, and TAPO on ASAP and ASAP++ under five-fold cross-validation. Scores are averaged over traits for each prompt. Values in parentheses denote changes in Avg. QWK over the corresponding SFT baseline, and the best result under each backbone is shown in bold.
}
\label{tab:prompt_post_training_appendix}
\end{table*}

\begin{table*}[t]
\centering
\scriptsize
\setlength{\tabcolsep}{6pt}
\renewcommand{\arraystretch}{1.05}
\begin{adjustbox}{max width=\textwidth}
\begin{tabular}{l|cccccccc|c}
\toprule
& \multicolumn{8}{c|}{\textbf{Prompts}} & \\
\midrule
\textbf{Method} 
& \textbf{P1} 
& \textbf{P2} 
& \textbf{P3} 
& \textbf{P4} 
& \textbf{P5} 
& \textbf{P6} 
& \textbf{P7} 
& \textbf{P8} 
& \textbf{Avg.} \\
\midrule
HISK        
& 0.674 & 0.586 & 0.651 & 0.681 & 0.693 & 0.709 & 0.641 & 0.516 & 0.644 \\
MTL-BiLSTM  
& 0.670 & 0.611 & 0.647 & 0.708 & 0.704 & 0.712 & 0.684 & 0.581 & 0.665 \\
STL-LSTM    
& 0.690 & 0.622 & 0.663 & 0.729 & 0.719 & 0.753 & 0.704 & 0.592 & 0.684 \\
DualTrans   
& 0.712 & 0.671 & 0.690 & 0.760 & 0.714 & 0.740 & 0.748 & 0.620 & 0.707 \\
ArTS        
& 0.700 & 0.699 & 0.704 & 0.766 & 0.725 & 0.770 & 0.739 & 0.644 & 0.718 \\
T-MES       
& 0.728 & 0.684 & 0.702 & 0.771 & 0.726 & 0.754 & 0.755 & 0.629 & 0.719 \\
SaMRL       
& 0.702 & 0.711 & 0.708 & 0.766 & 0.722 & 0.773 & 0.743 & 0.649 & 0.722 \\
RMTS        
& 0.716 & 0.704 & \textbf{0.723} & \underline{0.772} & \textbf{0.737} & 0.769 & 0.736 & 0.651 & 0.726 \\
GAT-AES     
& 0.722 & 0.698 & 0.707 & 0.771 & 0.720 & 0.767 & 0.744 & \underline{0.680} & 0.726 \\
RaDME       
& 0.705 & 0.716 & \underline{0.715} & \underline{0.772} & \underline{0.731} & \underline{0.774} & \underline{0.762} & 0.654 & \underline{0.729} \\
\midrule
TAPO        
& \textbf{0.730} & \textbf{0.737} & 0.713 & \textbf{0.782} & 0.728 & \textbf{0.793} & \textbf{0.765} & \textbf{0.694} & \textbf{0.743} \\
\bottomrule
\end{tabular}
\end{adjustbox}
\caption{
Prompt-wise QWK results on ASAP and ASAP++. Each value is averaged over the traits annotated for the corresponding prompt. Baselines are ordered by average QWK. The best and second-best results among reported methods are shown in bold and underlined, respectively.
}
\label{tab:prompt_main_results}
\end{table*}

\begin{table*}[t]
\centering
\scriptsize
\setlength{\tabcolsep}{6pt}
\renewcommand{\arraystretch}{1.05}
\begin{adjustbox}{max width=\textwidth}
\begin{tabular}{l|cccccccc|c}
\toprule
& \multicolumn{8}{c|}{\textbf{Prompts}} & \\
\midrule
\textbf{Setting} 
& \textbf{P1} 
& \textbf{P2} 
& \textbf{P3} 
& \textbf{P4} 
& \textbf{P5} 
& \textbf{P6} 
& \textbf{P7} 
& \textbf{P8} 
& \textbf{AVG} \\
\midrule
Full model 
& 0.730 
& 0.737 
& \textbf{0.713} 
& \textbf{0.782} 
& \textbf{0.728} 
& \textbf{0.793} 
& 0.765 
& 0.694 
& \textbf{0.743} \\
\midrule
w/o trait reward
& \textbf{0.732} 
& 0.721 
& 0.707 
& 0.772 
& 0.711 
& 0.779 
& \textbf{0.779} 
& \textbf{0.703} 
& 0.738 \\
w/o $R_{\mathrm{relation}}$
& 0.725 
& \textbf{0.746} 
& 0.710 
& 0.780 
& 0.722 
& 0.778 
& 0.770 
& 0.694 
& 0.741 \\
w/o $R_{\mathrm{format}}$
& 0.719 
& 0.745 
& 0.709 
& 0.779 
& 0.724 
& 0.789 
& 0.765 
& 0.687 
& 0.740 \\
w/o enhanced prompt
& 0.720
& 0.717
& 0.701
& 0.768
& 0.723
& 0.773
& 0.759
& 0.687
& 0.731 \\
\bottomrule
\end{tabular}
\end{adjustbox}
\caption{
Prompt-wise ablation results on ASAP and ASAP++. Scores are averaged over traits for each prompt. Full model denotes the complete version of TAPO, and the best result in each column is shown in bold.
}
\label{tab:prompt_ablation}
\end{table*}

We use the Huber penalty to compute score-matching rewards. 
For a normalized prediction error \(e\), the Huber penalty is defined as
\begin{equation}
H_{\delta}(e)=
\begin{cases}
\frac{1}{2}e^2, & |e|\leq \delta,\\
\delta\left(|e|-\frac{1}{2}\delta\right), & |e|>\delta.
\end{cases}
\end{equation}
Scores are first normalized to \([0,1]\) according to the corresponding prompt-specific score range, and \(e\) denotes the difference between the normalized prediction and gold score. We set \(\delta=0.1\) for all experiments. Since the error is normalized, this threshold treats deviations within 10\% of the score range as small errors and provides smooth quadratic feedback for them, while larger errors are penalized approximately linearly. We keep \(\delta\) fixed as an empirical smoothing constant and focus our hyperparameter analysis on the reward weights that directly control the relative contributions of different reward components.

\section{Qwen3-1.7B Setting}
\label{app:qwen_implementation}

We use Qwen3-1.7B as a decoder-only causal language model with a chat-style input format consisting of a system prompt and a user message. The user message follows the enhanced input setting used in the main experiments, including the original prompt text and LLM-generated trait descriptions. During SFT, we apply LoRA with rank $r=64$, $\alpha=128$, and dropout 0.05 to the \texttt{q\_proj}, \texttt{k\_proj}, \texttt{v\_proj}, and \texttt{o\_proj} modules. We use a maximum input length of 2048 tokens, a maximum generation length of 128 tokens, batch size 8, gradient accumulation 2, learning rate $5\times10^{-5}$, warmup ratio 0.05, and train for up to 25 epochs with early stopping patience 3.

For TAPO training, we initialize from the best SFT checkpoint and use LoRA with rank $r=32$, $\alpha=64$, and dropout 0.05 on the \texttt{q\_proj} and \texttt{v\_proj} modules. We use group size 4, batch size 1, gradient accumulation 8, learning rate $2\times10^{-5}$, KL coefficient $\beta_{\mathrm{KL}}=0.1$, and train for 2 epochs with 3 evaluations per epoch. Generation uses temperature 0.7, top-$k=50$, top-$p=0.95$, and a maximum of 128 new tokens. The reward components follow the T5-large setting, with token-level shaping applied to trait-specific score tokens, token reward weight 0.3, and maximum token advantage clipped to 10.0.

\section{Prompt-wise results}
\label{app:prompt_results}

This section provides prompt-wise results corresponding to the trait-wise analyses in the main text. Table~\ref{tab:prompt_post_training_appendix} compares SFT, GRPO, and TAPO under T5-large and Qwen3-1.7B. TAPO achieves the best average prompt-wise QWK under both backbones, improving over GRPO from 0.738 to 0.743 on T5-large and from 0.715 to 0.726 on Qwen3-1.7B.

Table~\ref{tab:prompt_main_results} compares TAPO with prior methods on ASAP and ASAP++. TAPO obtains the best average QWK of 0.743 and achieves the best result on 5 out of 8 prompts. Table~\ref{tab:prompt_ablation} further reports prompt-wise ablation results, where removing each major component consistently lowers the average performance. These findings are consistent with the trait-wise results in the main text.

\section{Standard Deviation Results}
\label{app:sd_results}

Table~\ref{tab:sd_results} summarizes the average QWK and five-fold standard deviation for methods that explicitly report standard deviations on ASAP and ASAP++. TAPO achieves the highest average QWK under both trait-wise and prompt-wise evaluations. Its standard deviations are comparable to those of prior autoregressive and structured baselines such as ArTS, SaMRL, and GAT-AES, and are substantially smaller than those reported for RMTS. These results indicate that TAPO improves average performance without introducing large cross-fold variance.

\begin{table}[t]
\centering
\small
\setlength{\tabcolsep}{6pt}
\renewcommand{\arraystretch}{1.05}
\begin{tabular}{lcc}
\toprule
\textbf{Method} & \textbf{Trait Avg.} & \textbf{Prompt Avg.} \\
\midrule
ArTS    & 0.702 ($\pm$0.013) & 0.718 ($\pm$0.012) \\
SaMRL   & 0.705 ($\pm$0.013) & 0.722 ($\pm$0.012) \\
RMTS    & 0.708 ($\pm$0.043) & 0.726 ($\pm$0.042) \\
GAT-AES & 0.710 ($\pm$0.011) & 0.726 ($\pm$0.009) \\
TAPO    & \textbf{0.726} ($\pm$0.012) & \textbf{0.743} ($\pm$0.013) \\
\bottomrule
\end{tabular}
\caption{
Average QWK and five-fold standard deviation on ASAP and ASAP++. We include only methods for which standard deviations are explicitly reported.
}
\label{tab:sd_results}
\end{table}

\section{Additional Implementation Details}
\label{app:additional_hyperparameter}

\paragraph{Hyperparameter settings.}
We further analyze the relation reward weight $\lambda_1$ and the format
reward weight $\lambda_2$. For $\lambda_1$, we compare three values, 0.1,
0.2, and 0.3, while keeping the other reward weights fixed. As shown in
Table~\ref{tab:relation_weight_analysis}, $\lambda_1=0.2$ achieves the best
performance under both prompt-wise and trait-wise evaluation. A smaller value
provides weaker inter-trait consistency guidance, while a larger value may
overemphasize pairwise ranking constraints and slightly hurt direct score
matching.

For the format reward, we fix $\lambda_2=0.1$ as a small auxiliary constraint
for structured output validity. Since this reward is used to encourage
parseable score generation, we use 0.1 as an empirical regularization weight
throughout the experiments.

\begin{table}[t]
\centering
\small
\setlength{\tabcolsep}{8pt}
\renewcommand{\arraystretch}{1.05}
\begin{tabular}{ccc}
\toprule
$\lambda_{1}$ & \textbf{Prompt Avg.} & \textbf{Trait Avg.} \\
\midrule
0.1 & 0.736 & 0.719 \\
0.2 & \textbf{0.743} & \textbf{0.726} \\
0.3 & 0.740 & 0.723 \\
\bottomrule
\end{tabular}
\caption{
Effect of different relation reward weights $\lambda_1$. Prompt Avg. and
Trait Avg. denote the average QWK under prompt-wise and trait-wise evaluations,
respectively.
}
\label{tab:relation_weight_analysis}
\end{table}

\paragraph{Implementation details.}
We locate score spans by applying regular-expression matching to the decoded
structured output and map the matched character spans back to their
corresponding token indices. Since the score values in our datasets are
integers, they are almost always represented by a single token. In our
implementation, the normalized trait-level local reward is assigned to the
ending token of each matched score span, while non-score tokens receive no
additional local reward.

If an output is malformed or cannot be reliably parsed, we set
$R_i^{\mathrm{fmt}}=-1$ and do not compute the remaining score-based reward
components for that output. For group-wise normalization, we include
$\epsilon$ in the denominator for numerical stability. When all sampled
outputs receive the same reward for a component, mean centering yields zero
normalized advantages, so that component contributes no gradient signal for
the group.

For the per-trait MTL+RL baseline in Section~\ref{sec:per_trait}, each GRPO
group consists of multiple rollouts for the same essay-trait pair, and the
resulting sample-level advantage is shared across all output tokens. For the
consistency-enhanced variant, we use the same range-normalized pairwise hinge
formulation as the relation reward in TAPO, but recompute it across
independently generated trait predictions for the same essay and add it to
the trait-specific reward with weight $\lambda_{\mathrm{cons}}$.

\section{Additional Statistical Analyses}
\label{sec:additional_analysis}

\subsection{Multi-Seed Comparison with Standard GRPO}
\label{sec:multiseed}

The main experiments use five-fold cross-validation, which primarily captures
variation across data splits. Since GRPO additionally involves stochastic
rollout sampling, we further repeat TAPO and standard GRPO with three random
seeds under identical enhanced prompts and structured output formats.
Table~\ref{tab:multiseed_performance} reports the resulting average
performance.

Although the absolute improvements are modest, TAPO consistently outperforms
standard GRPO across the evaluated dimensions. We further perform paired
tests over corresponding prompt-trait cells, traits, and prompts.
As shown in Table~\ref{tab:multiseed_significance}, TAPO improves on
40 of 44 prompt-trait cells, all 11 traits, and all 8 prompts.

\begin{table}[t]
\centering
\scriptsize
\setlength{\tabcolsep}{3.2pt}
\renewcommand{\arraystretch}{1.05}
\begin{adjustbox}{max width=\columnwidth}
\begin{tabular}{l|ccc}
\toprule
\textbf{Method}
& \textbf{Prompt Avg.}
& \textbf{Trait Avg.}
& \textbf{Mean} \\
\midrule
Standard GRPO
& 0.7350
& 0.7193
& 0.7272 \\
TAPO
& \textbf{0.7392}
& \textbf{0.7239}
& \textbf{0.7316} \\
\midrule
$\Delta$
& +0.0042
& +0.0046
& +0.0044 \\
\bottomrule
\end{tabular}
\end{adjustbox}
\caption{
Three-seed comparison between standard GRPO and TAPO under identical
enhanced prompts. Prompt Avg. and Trait Avg. denote prompt-wise and
trait-wise QWK, respectively.
}
\label{tab:multiseed_performance}
\end{table}

\begin{table}[t]
\centering
\scriptsize
\setlength{\tabcolsep}{2.5pt}
\renewcommand{\arraystretch}{1.05}
\begin{adjustbox}{max width=\columnwidth}
\begin{tabular}{l|c|c|c|c|c}
\toprule
\textbf{Unit}
& $\mathbf{N}$
& \textbf{Diff.}
& \textbf{W/L/T}
& $\mathbf{p_t}$
& $\mathbf{p_w}$ \\
\midrule
Prompt-trait
& 44
& +0.0044
& 40/4/0
& $<0.0001$
& $<0.0001$ \\
Trait
& 11
& +0.0046
& 11/0/0
& $<0.0001$
& 0.0005 \\
Prompt
& 8
& +0.0042
& 8/0/0
& 0.0023
& 0.0039 \\
\bottomrule
\end{tabular}
\end{adjustbox}
\caption{
One-sided paired significance analysis between TAPO and standard GRPO,
using TAPO $>$ GRPO as the directional alternative.
Diff. denotes the mean paired QWK difference (TAPO minus GRPO), and
W/L/T denotes the numbers of wins, losses, and ties for TAPO.
$p_t$ and $p_w$ denote the paired $t$-test and Wilcoxon signed-rank
test, respectively.
}
\label{tab:multiseed_significance}
\end{table}

These results indicate that the improvement over standard GRPO is
consistent across random seeds and evaluation granularities, although
the absolute gain remains relatively small.

\subsection{Additional Comparison with RaDME}
\label{sec:radme_analysis}

We additionally examine the consistency of TAPO's improvement over RaDME,
the strongest prior baseline in our main comparison, and assess how much of
this advantage remains without enhanced prompting. Since fold-level or
seed-level predictions from RaDME are unavailable, a directly matched
fold-level statistical comparison is not possible. We therefore pair the
reported results across the 11 trait dimensions and 8 prompt dimensions,
respectively, for both the full TAPO model and TAPO without enhanced prompts.

As shown in Table~\ref{tab:radme_analysis}, full TAPO obtains mean gains
of 0.0150 and 0.0141 over RaDME across traits and prompts, respectively,
outperforming RaDME on 9 of 11 traits and 6 of 8 prompts. Both improvements
are significant under the paired $t$-test and Wilcoxon signed-rank test.
When the enhanced prompt is removed, the gains decrease to 0.0052 across
traits and 0.0024 across prompts, and neither comparison remains
statistically significant. These results indicate that enhanced prompting
contributes substantially to TAPO's advantage over RaDME, while the remaining
TAPO formulation remains competitive with the prior baseline. We emphasize
that these analyses pair aggregate trait- and prompt-level results rather
than matched folds or random-seed runs and should therefore be interpreted
accordingly.

\begin{table}[t]
\centering
\scriptsize
\setlength{\tabcolsep}{2.2pt}
\renewcommand{\arraystretch}{1.05}
\begin{adjustbox}{max width=\columnwidth}
\begin{tabular}{l|l|c|c|c|c|c}
\toprule
\textbf{Setting}
& \textbf{Unit}
& $\mathbf{N}$
& \textbf{Gain}
& \textbf{W/L}
& $\mathbf{p_t}$
& $\mathbf{p_w}$ \\
\midrule
\multirow{2}{*}{TAPO}
& Trait
& 11
& +0.0150
& 9/2
& 0.0014
& 0.0024 \\
& Prompt
& 8
& +0.0141
& 6/2
& 0.0157
& 0.0234 \\
\midrule
\multirow{2}{*}{w/o enhanced prompt}
& Trait
& 11
& +0.0052
& 8/3
& 0.0638
& 0.0703 \\
& Prompt
& 8
& +0.0024
& 3/5
& 0.3330
& 0.5938 \\
\bottomrule
\end{tabular}
\end{adjustbox}
\caption{
One-sided paired comparisons against RaDME over the reported trait-wise
and prompt-wise results, using the corresponding TAPO variant $>$ RaDME
as the directional alternative. Gain denotes the mean QWK difference
relative to RaDME. $p_t$ and $p_w$ denote the paired $t$-test and
Wilcoxon signed-rank test, respectively.
}
\label{tab:radme_analysis}
\end{table}

\section{Enhanced Prompt Details}
\label{app:enhanced_prompt}

We provide the prompt texts and trait descriptions used in the enhanced input. 
The enhanced input supplements the prompt ID with the original writing prompt and descriptions of the applicable scoring traits.

\begin{appendixpromptbox}{Prompt Texts}
\begin{enumerate}[leftmargin=*, label=\textbf{P\arabic*:}, itemsep=2pt, topsep=2pt]
    \item Write a letter to your local newspaper in which you state your opinion on the effects computers have on people.
    
    \item Write a persuasive essay to a newspaper reflecting your views on censorship in libraries.
    
    \item Write a response that explains how the features of the setting affect the cyclist. In your response, include examples from the essay that support your conclusion.
    
    \item Read the following passage: ``When they come back, Saeng vowed silently to herself, in the spring, when the snows melt and the geese return and this hibiscus is budding, then I will take that test again.'' Write a response that explains why the author concludes the story with this paragraph. In your response, include details and examples from the story that support your ideas.
    
    \item Describe the mood created by the author in the memoir. Support your answer with relevant and specific information from the memoir.
    
    \item Based on the excerpt, describe the obstacles the builders of the Empire State Building faced in attempting to allow dirigibles to dock there. Support your answer with relevant and specific information from the excerpt.
    
    \item Do one of the following: write a story about a time when you were patient OR write a story about a time when someone you know was patient OR write a story in your own way about patience.
    
    \item Tell a true story in which laughter was one element or part.
\end{enumerate}
\end{appendixpromptbox}

\begin{appendixpromptbox}{Trait Descriptions}
\textbf{Content:} Assess the development of ideas, quality of arguments, relevance of details, and clarity of the message.

\textbf{Organization:} Evaluate the logical structure, effectiveness of the introduction and conclusion, and use of transitional phrases.

\textbf{Word Choice:} Rate the precision, variety, and appropriateness of vocabulary used to convey meaning.

\textbf{Sentence Fluency:} Judge the rhythm, flow, and variety of sentence structures. High scores sound natural when read aloud.

\textbf{Conventions:} Check for correctness in grammar, spelling, punctuation, capitalization, and paragraphing.

\textbf{Voice:} Evaluate the individual personality, tone, and engagement with the audience. The writing should feel authentic.

\textbf{Prompt Adherence:} Critically assess whether the essay answers all parts of the prompt and references the source text as required.

\textbf{Language:} Evaluate the overall command of English, including grammar solidity and vocabulary usage.

\textbf{Narrativity:} Assess the cohesion and fluidity of the text. High scores are written as a unified essay, not a disjointed list of answers.

\textbf{Style:} Evaluate the writer's command of tone, voice, and rhetorical devices to engage the reader.

\textbf{Overall:} The holistic quality of the essay, considering how well all traits combine to achieve the writing purpose.
\end{appendixpromptbox}

\end{document}